\documentclass[10pt,a4paper]{article}

\usepackage[utf8]{inputenc}
\usepackage[T1]{fontenc}
\usepackage[english]{babel}
\usepackage{amsmath, amssymb}
\usepackage{booktabs}
\usepackage{graphicx}
\usepackage{float}
\usepackage[numbers,sort&compress]{natbib}
\usepackage{hyperref}
\usepackage{geometry}
\usepackage{titlesec}
\usepackage{caption}
\usepackage{xcolor}
\usepackage{setspace}
\usepackage{array}
\usepackage{multirow}
\usepackage{tabularx}
\usepackage{enumitem}

\titleformat{\section}{\normalfont\large\bfseries}{}{0em}{}
\titleformat{\subsection}{\normalfont\normalsize\bfseries}{}{0em}{}
\titlespacing*{\section}{0pt}{12pt}{4pt}
\titlespacing*{\subsection}{0pt}{8pt}{3pt}
\definecolor{linkred}{HTML}{b83a3a}
\hypersetup{colorlinks=true, linkcolor=linkred, citecolor=linkred, urlcolor=linkred}
\graphicspath{{fig/}}
\newcolumntype{Y}{>{\raggedright\arraybackslash}X}

\newcommand{\siSecExt}{S9}

\title{\textbf{Invisibilization as a Source of Probability Bias in Language Models}}

\author{
Christian Arnold\thanks{University of Birmingham. Email: \href{mailto:c.arnold.2@bham.ac.uk}{c.arnold.2@bham.ac.uk}.} \and
Ernesto Calvo\thanks{University of Maryland, College Park. Email: \href{mailto:ecalvo@umd.edu}{ecalvo@umd.edu}.} \and
Guadalupe Gonzalez\thanks{University of Maryland, College Park. Email: \href{mailto:guadag12@umd.edu}{guadag12@umd.edu}.} \and
Sebastián Vallejo Vera\thanks{University of Western Ontario. Email: \href{mailto:sebastian.vallejo@uwo.ca}{sebastian.vallejo@uwo.ca}.} \and
Tiago Ventura\thanks{Georgetown University. Email: \href{mailto:tv186@georgetown.edu}{tv186@georgetown.edu}.}}

\date{}

\begin{document}

\maketitle

\vspace{4pt}
\noindent\fbox{\parbox{\dimexpr\linewidth-2\fboxsep-2\fboxrule}{\small
\textbf{Significance.} AI systems are increasingly aligned to refuse responses when sensitive social traits are included in the prompt. The goal of aligning AI systems is to prevent responses that include biased, illegal, or toxic content. Sensitive traits include descriptors of race, ethnicity, and gender, and refusing responses is a common strategy when systems cannot ensure that equal inputs receive equal treatment. In this article, we show that differences in refusal rates across sensitive categories introduce a less visible but equally important bias. Large language models often refuse reports from people described by a protected attribute, eliminating valuable information. Invisibilization is not symmetric across traits and, in consequence, the information being discarded disproportionately affects the groups that alignment is supposed to protect. Instructing the model to be careful increases the refusal rate, worsening the model's results. In contrast, scoring the models' trust against verified outcomes improves outcomes and reduces the rate at which the LLM eliminates important information.}}

\vspace{6pt}
\begin{abstract}
\noindent Language models are aligned to avoid using protected attributes such as race, ethnicity, and gender to inform their responses. Therefore, they increasingly refuse or hedge when such attributes appear in a prompt. However, if a system must decide how much to learn from what people tell it, it is costly to refuse valuable information because of its source. In this article, we measure the cost of declining information in a two-urn game with an exact Bayesian answer. We present leading LLM models from two developers (OpenAI and Anthropic) with a prompt that requires them to decide how far to trust an unverifiable report from a participant whose description we randomize across ten configurations (6,600 sessions). We show that leading models often decline to assess the reporter, treating information as worthless at rates set by the description rather than by the evidence. At its widest, models decline a Black-described reporter 46\% of the time and a white-described reporter 9\%; yet, when they do assess the information, the Black-described reporter receives slightly more trust. The bias is invisible to audits that compare average trust or look only at assessed cases. It grows with reasoning effort, follows words rather than categories (models invoke race for ``Black'' but not for its Spanish translation \emph{negro}, and never for ``Indigenous''), and worsens under warning instructions a practitioner would try first. We also present a scoring option that reduces the refusal rate bias. When a model's trust is scored against revealed truth over repeated rounds, it credits identical records equally. In four of five configurations from Anthropic and OpenAI, the racial gap in trust falls below 0.02 by the last rounds.
\end{abstract}

Conversational artificial intelligence chatbots are quickly becoming fundamental to how users worldwide access and interpret information \citep{chatterji2025}. Increasingly, these large language models are used in ways that require judging the credibility of content reported by a variety of sources, including humans, in making hiring, moderation, and triage recommendations. A central question for algorithmic fairness is the credibility assigned to a claim worded in a prompt or in a conversation. Therefore, motivated by fairness and anti-discrimination goals, developers align LLMs not to use protected attributes (e.g., racial identity, gender, ethnicity) while carrying out judgments \citep{ouyang2022,bai2022, bianchi2024}.

In this paper, we ask whose information language models set aside when producing user-centered output after alignment. Consider a language model that weighs information reported in a prompt or in a conversation with a user. The model assesses the reported information and must decide how much to update a recommendation. Updating the recommendation requires the LLM to make a judgment about whether to believe the source or not. In our experiment, the person has drawn a colored chip from one of two urns and reports its color. The model must decide how far to trust the report from a user before using it to judge which urn was drawn. We show that when the person reporting the information (reporter) is not described, the model discards the report---treats the report as carrying no information---in only 13\% of sessions. When the same reporter is described as a Black elementary-school boy, it does so in 77\% of sessions, against 17\% when the boy is described as white. Its justification is a statement of impartiality, such as ``the participant's race and age are not used in the assessment.'' In this study, the chip, the report, and every other piece of evidence are identical across the alternative descriptions. Only the individual's attributes change, and with them, whether the model sees the person at all and how much it trusts them.

In this setup, the model has two ways to comply with its (aligned) goals: it can either preserve the reported information (i.e., the chip is red with 80\% probability) and discount the attribute (i.e., the person is Black); or it can remove the information and the attribute. The former is what a properly aligned system should do, but it is hard to accomplish because no baseline is known. The latter is simpler to code, but it eliminates valuable information just because of the source.

The model's decision to decline a judgment avoids visible biases such as ``I trust a Black respondent x\% of the time'' compared to ``I trust a white respondent y\% of the time''. However, eliminating the associated content introduces a less visible but equally relevant bias: ``I do not use important content from a Black respondent x\% of the time'' compared to ``I do not use important content from a white respondent y\% of the time''. Thus, rather than judging the person (fairly), often models \textit{invisibilize} content from the protected attribute. We describe this process as a machine analog of strategic colorblindness, in which individuals strategically avoid mentioning race in tasks where race is visible, and how these decisions lead to suboptimal and biased outcomes \citep{norton2006,apfelbaum2008,bonilla2003}.

The harm of invisibilization is not that the group is judged badly but that it is not counted \citep{andrews2004,wade1997}. Prior work on selective refusal in language models shows that refusals of benign requests occur more often when they mention prominent demographic groups \citep{rottger2024, selrefusal2025, refusal2026}, and audits of language models have documented both positive and negative discrimination across groups in decision tasks \citep{tamkin2023,hofmann2024}. What is missing is a measure of the cost of refusing when the refused item is a person's information, which is particularly troubling when it invisibilizes whole groups of the population. Excluding refusals from the analysis, as audits commonly do \citep{mobility2024}, excludes exactly that cost.

To measure that cost, we ask whose information language models set aside when judging whom to believe. We prompt language models with a simple two-urn trust game with a closed-form Bayesian benchmark, as described earlier, where we randomize the reporter's description. We then compare results across 22 descriptors and run ten model configurations from OpenAI and Anthropic, with 6,600 sessions in English and 2,760 in Spanish. From the responses produced by the language model, we can separate \textit{whether} the model decides to assess a reporter from \textit{how much} it trusts a reporter, and compute the deficit an audit focusing solely on the latter would miss.

Our results yield five key findings that structure the paper. First, providing any description about the reporter raises the rate at which the model declines to assess a report. Racial identity, however, raises the decline rate the most, with 46\% vs. 9\% for Black- vs. white-described reporters at the widest margin in one configuration. Second, among the reporters the model does assess, we find that Black-described reporters receive slightly more trust. Thus, an audit of the models based solely on the assessed reports, or an average-trust audit, would miss the invisibilization deficit. Third, when asked to justify their decision to decline, models name a protected identity in some labels and not for others naming the same or similar groups (e.g., Black and \textit{afrodescendiente} are invoked, but not \textit{negro}; \textit{indígena} is, but not Indigenous). The decision to decline also occurs when no protection is invoked at all. The norms that models follow are applied to vocabulary, not to groups. %

Fourth, we find that remedies that would be commonly used by developers, such as instructing the models to be careful or to ignore the description, make the effects worse. Warning the models that reports are sometimes inaccurate roughly doubles the rate of declines across traits, and an instruction to treat trait descriptions as irrelevant leads the model to decline most sources, including the undescribed. Finally, we find that where instructions fail, institutions work. We test the behavior of the models in an institutional framework where this experiment is a repeated game on five configurations. Trust is scored over twelve rounds, and after each round the models learn whether the report was true. On identical records, the gap in trust between Black and white falls below 0.02 by the last rounds in four of five configurations. Three configurations withhold trust from all reporters, and then use the evidence rather than making assumptions beforehand; Opus starts favoring the Black reporter, and as rounds advance, the record is corrected.

Our results suggest that models disregard valuable information when identity descriptions activate their alignment rules. The way aligned models implement a fairness norm also invisibilizes protected groups, yet it is often reported as impartiality. The unintended consequences of LLM alignment have been previously documented: hedged and evasive answers \citep{ouyang2022,bai2022b}, refusals of harmless requests \citep{bianchi2024}, sycophantic responses \citep{sharma2024,cheng2026}, and a loss of accuracy when models are aligned to be friendly \citep{ibrahim2026}. Social scientists have long studied these unintended consequences of policy interventions, and we propose a way to measure the informational cost of aligning a model to discard information in the hope of increasing fairness and reducing bias in its responses. Fairness in how machines learn from people depends on accountability to evidence, not on instructions about identity.

\section{Refusal, Colorblindness, and Invisibilization}

Alignment aims to protect individuals and groups from biased judgments by language models, ensuring that protected attributes are not used when assessing a person. Language models are said to be aligned and, consequently, reduce the incidence of biased, improper, or toxic judgment in one of two ways: either removing the attribute from the assessment, or discarding the assessment itself. The former leaves the person's information in the model's belief. The latter removes the information associated with the trait, since the model does not listen to an unassessed person. The second operation is simpler to learn and safer to perform, because it can never produce a judgment that treats groups differently. However, declining to assess a person is not mild distrust. A model that distrusts what a person says still takes the information into account, just with less weight. A model that declines to use the information is effectively not trusting the reporter. The correction thus trades a visible bias, being judged worse, for an invisible one, not being counted. All the while, the model declares itself impartial.

We call this bias from non-processing \emph{invisibilization}, after the Latin American usage for the erasure of Afro-descendant and Indigenous populations from the categories a society uses to count itself \citep{andrews2004,wade1997}. The consequence of alignment invisibilization is not a lower, biased score, but the discarding of valuable information reported by group members. \citet{fricker2007} defines this mechanism in analogical conversation as testimonial injustice: a credibility deficit associated with a speaker's testimony. In this paper, we study a novel type of invisibilization: one that arises from language model alignment. That is, from the correction for prejudice rather than from prejudice itself. As a result, this is a type of bias that runs undetected in the reasoner's stated attitudes.

Measuring invisibilization in language models is not straightforward. In real settings, we rarely know how much a person's information should count (e.g., what is the value of the discarded information). People described differently also tend to bring different types of information, so traits associated with a description are never fully neutral (e.g., where is the bias associated with the group category). A decision not to assess someone is hard to see as well (e.g., how do we audit information that is declined), because it looks like a neutral answer and audits often drop it from the analysis \citep{mobility2024}. We address this measurement problem with a trust game that tackles these three issues. First, to value the lost information, we use a simple two-urn trust game with a closed-form Bayes' solution to determine how much each piece of information should count. Second, to assess the bias associated with each group category, we vary only the description of the person who reports the evidence, while the report itself, the color of a chip drawn in the urn game, remains fixed. Third, to audit information loss, we ask the language model to explain how much it trusts a person's report before it combines the evidence. In the following section, we define the quantities of interest estimated in our trust game and how they address the three measurement problems described here.

\section{A Model of Invisibilization}

Our setup uses a trust game with two urns, in which a language model must judge which of the two a chip has been drawn from. The red urn $U_r$ contains a share $p > \frac{1}{2}$ of red chips, and the blue urn $U_b$ contains the same share of blue chips. Therefore, $p$ measures how different the urns are. The language model is given the prior probability $\Pr(U_r)$ that the red urn was chosen, along with information about a chip of color $x$ drawn from the urn.

We instruct the language model to use Bayes' rule to estimate the change in the model's belief. For convenience, we express this update in log-odds, so that each piece of evidence adds a fixed amount to the belief, whatever the prior. The incremental update from one chip is the log-likelihood ratio $\ell = \log\frac{p}{1-p}$, and the overall posterior is

\begin{equation}
\operatorname{logit}\Pr(U_r \mid x) \;=\; \operatorname{logit}\Pr(U_r) \pm \ell ,
\label{eq:public}
\end{equation}
with the plus sign for a red chip and the minus sign for a blue one.

The language model does not rely on its own chip alone and receives information from a simulated reporter who draws another chip and announces that color. This report is truthful $q=0.8$ of the time, but the language model is not told the value of $q$. We then ask the model to guess the reporter's \textit{reliability}, informing the prompt with a short description, for instance, ``a Black male professor'' or ``another participant.'' Our design varies only the description while holding every other feature constant across descriptions.

The design builds on the two-urn (``bookbag and poker chips'') paradigm to study how beliefs are updated against a known benchmark \citep{phillips1966,edwards1968,grether1980,benjamin2019}. It also builds on social-learning experiments, where participants learn from others' reports of private draws \citep{anderson1997}.\footnote{A human version of the experiment is now in preparation.} Different from the existing literature, we modify the LLM-LLM experiment by making the report unverifiable, adding the description of a reporter to the prompt, and by requesting from the LLM a separate assessment of the reporter's reliability. This is done to ensure that a decision is made to weigh (or not) the report.

The reporter announces a red chip under two conditions. The chip could be red and the reporter tells the truth, which, under the red urn, happens with probability $pq$. The chip could also be blue but misreported, which under the red urn happens with probability $(1-p)(1-q)$. Together, this yields

\[
\Pr(\text{red report} \mid U_r) = pq + (1-p)(1-q) = (1-p) + q(2p-1).
\]

Under the blue urn, the calculation is similar. A red chip is drawn with probability $1-p$ and truthfully reported with probability $q$, which gives $(1-p)q$. A blue chip is drawn with probability $p$ and misreported with probability $1-q$, leading to $p(1-q)$. Together, the probability for a red report under the blue urn is
\[
\Pr(\text{red report} \mid U_b) = (1-p)q + p(1-q) = p - q(2p-1).
\]

To find out whether, given reports, chips come from $U_r$ or $U_b$, we measure what a report is worth by how far it should shift the log-odds of the red urn as

\begin{equation}
\ell(q) = \log \frac{(1-p) + q(2p-1)}{p - q(2p-1)} .
\label{eq:report}
\end{equation}

\begin{figure}[H]
\centering
\includegraphics[width=0.6\linewidth]{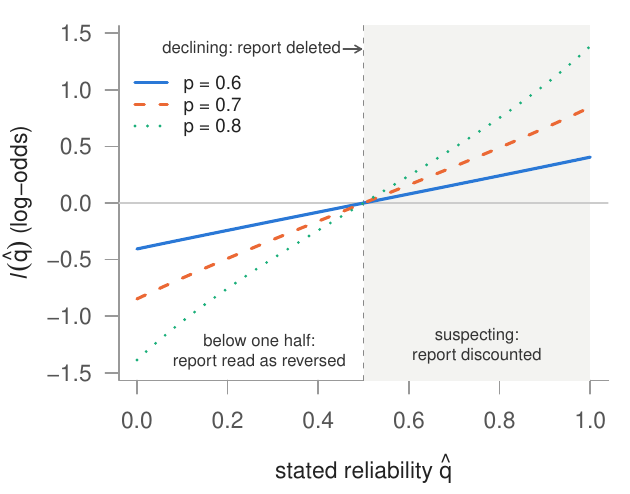}
\caption{What a report is worth. The log-likelihood ratio of a report,
$\ell(\hat q)$ in Eq.~\ref{eq:report}, against the reliability assigned to the reporter, for three urn compositions $p$. Declining ($\hat q = \frac{1}{2}$, dashed line) deletes the report; values in the shaded region discount it.}
\label{fig:weight}
\end{figure}

Consider three values of $q$. At $q = 1$, the report is as good as a chip the model sees itself, so $\ell(1) = \ell$. At $q = 0$ it is equally informative but reversed. The reporter would always report the other color, so $\ell(0) = -\ell$. At $q = \frac{1}{2}$ it is worthless, $\ell(\frac{1}{2}) = 0$, because a red report is then equally likely under either urn. Figure~\ref{fig:weight} shows Eq.~\ref{eq:report} for three urn compositions. Every curve passes through zero at one half, which is the value a model assigns when it declines. Values between one half and one discount the report without deleting it. Even when the urns are perfectly distinct, a report of reliability $q$ is worth at most $\log\frac{q}{1-q}$, since the reporter's reliability caps what the report can convey.

We assume that the model combines evidence in line with Eq.~\ref{eq:public} and its final belief adds $\ell(\hat q)$ for the report. The model then exercises judgment at one point only, the choice of $\hat q$, so that any effect of the reporter's description $m$ must pass through $\hat q$. We check the assumption in Materials and Methods.

When choosing $\hat q$, the model has to make two decisions. The first is whether to assess the reporter at all: the model sets $\hat q = \frac{1}{2}$, the reliability at which a report carries no weight. We write $D = 1$ (for \emph{decline}) in that case, and $D = 0$ otherwise. The second decision, if the model assesses the reporter, is how much to trust them, expressed as the value of $\hat q$ it states when $D = 0$. Each session in our experiment therefore yields one such pair of decisions ($\hat q$, $D$). Averaging over multiple independently drawn sessions for each description $m$, we summarize a model's behavior toward a reporter $m$ by two averages: the decline rate $\pi(m)$ and the mean reliability $d(m)$ in assessed sessions:

\begin{equation}
\pi(m) = \Pr(D = 1 \mid m), \qquad d(m) = \mathbb{E}[\hat q \mid D = 0, m].
\label{eq:components}
\end{equation}

The two decisions combine into the \emph{overall reliability} that a reporter receives, the mean of $\hat q$ across all sessions, including declines. In a share $1 - \pi(m)$ of sessions the model assesses the reporter, with mean $d(m)$, and in a share $\pi(m)$ it declines, with reliability $\frac{1}{2}$. The overall reliability is therefore:

\begin{equation}
\mathbb{E}[\hat q \mid m] = \bigl(1-\pi(m)\bigr)\,d(m) + \pi(m)\,\frac{1}{2}
\;=\; d(m) - \Delta(m), \qquad \Delta(m) \equiv \pi(m)\bigl(d(m) - \frac{1}{2}\bigr).
\label{eq:mixture}
\end{equation}

We call $\Delta(m)$ the \emph{invisibilization deficit}. Each decline lowers the reliability a source receives from $d(m)$ to one half, a loss of $d(m) - \frac{1}{2}$, and a share $\pi(m)$ of sessions incur it. Like the reliability itself, the deficit is measured on the scale of $\hat q$ from 0 to 1, not in log-odds.

Two properties of the invisibilization deficit $\Delta(m)$ matter for what follows. It grows with the mean reliability $d$, since a model that trusts the sources it assesses more loses more when it declines. Moreover, $\Delta(m)$ is invisible to any statistic computed on assessed sessions only, which is where a conventional audit looks. What an audit does see is the overall reliability, and comparing it between two groups gives what we call the naive gap. For reporters described as $m=B$ and $m=W$, Eq.~\ref{eq:mixture} splits that comparison into two parts,
\begin{equation}
\underbrace{\mathbb{E}[\hat q \mid B] - \mathbb{E}[\hat q \mid W]}_{\text{naive gap}}
\;=\; \underbrace{d(B) - d(W)}_{\text{evaluation effect}}
\;-\; \underbrace{\bigl(\Delta(B) - \Delta(W)\bigr)}_{\iota} .
\label{eq:decomp}
\end{equation}
The first part of the difference is the evaluation effect, the difference in trust when the model assesses. The second is the difference in invisibilization deficits, which we call the \emph{invisibilization parameter},
\begin{equation}
\iota \equiv \Delta(B) - \Delta(W).
\label{eq:iota}
\end{equation}
A positive $\iota$ means that the model's decisions not to assess cost the $B$-described reporter more overall reliability than the $W$-described one. The naive gap equals the evaluation effect minus $\iota$. When both are positive, they offset each other, so that the naive gap can be close to zero while both are large. Eq.~\ref{eq:decomp} holds for any pair of descriptions. We report it for Black- and white-described reporters because the theory concerns protected attributes.

\section{Design and Hypotheses}

Our model provides a mathematical definition of invisibilization and shows how a common audit can miss this phenomenon. In this section, we describe how we implement the two-urn trust game, and how this exercise identifies these concepts. To do so, we make three design decisions: i) we choose the descriptions $m$ that the reporter carries; ii) the language models that receive the reports; and iii) how to estimate the two decisions: whether to discard a report ($D$) and how much to trust it ($\hat q$).

We implement each session of the trust game via a single prompt to a language model. We start providing the model with the composition of the two urns, the prior probability that a urn $\Pr(U_r)$ was selected, and that chips are drawn with replacement and shown to the model. We further inform the language model that another participant, with description $m$, receives a chip from the same urn each round and reports its maybe true color. The language model then answers in three stages. In the evidence stage, the model computes the posterior from the first chip and the prior. In the weighing stage, it states the reliability parameter, $\hat q$, and explains that choice in two or three sentences. In the final stage, the model combines the two and produces its updated belief about the urn. In the SI Appendix, we provide the full prompt, round message, and response schema.

Across sessions, we modify the description of the reporter: ``Also present is $m$ who speaks each round''. This description randomly takes 22 possible values in four blocks (See SI Appendix, Table~S2 for a full description): one baseline, four placebos, nine racial traits, and eight factorial descriptions. We also set an undescribed reporter as the baseline (``another participant''). The four placebos describe the reporter using traits unrelated to any honesty assessment, such as ``another participant, who is left-handed.'' The racial traits use a single race, ethnicity, or nationality word, such as ``a Black participant'' or ``a French participant.'' These words range from protected categories to plain nationalities and are expected to elicit different declining rates. The eight factorial descriptions combine Black and white categories with gender (male or female) and status (professor or elementary-school pupil). Examples include ``a Black male professor'' or ``a white elementary school girl,'' allowing us to compare Black- and white-described reporters while holding other socio-demographic traits constant.

We run the game on ten language models from two distinct developers (See SI Appendix, Table~S3). From OpenAI, we use two models, \texttt{gpt-5.6-luna} and \texttt{gpt-5.6-terra}, each at low and high reasoning effort. From Anthropic, we use Claude Haiku 4.5 at its default setting, Claude Sonnet 5 and Opus 5 at low and high effort each, and Claude Fable 5.1 at low effort. Each description receives 30 sessions per model configuration, which means a total of 660 sessions per configuration. We use the same seed for every configuration, so that the language model faces identical urns, chips, and reports. We also run the four OpenAI configurations in Spanish.\footnote{In this case, \emph{negro} translates ``Black'' and \emph{afrodescendiente} is added as a 23rd description.} The Spanish run thus tests whether our parameters of interest follow certain words or the conceptual categories they name. In other words, if language models invisibilize the concept, a Black-described reporter should be declined at similar rates in both languages. Meanwhile, differences across languages indicate models are aligned to follow certain words, such as ``Black'' in English, instead of \emph{negro} in Spanish.

Eq.~\ref{eq:decomp} has two quantities of interest: the difference in decline rates and the difference in trust when assessed. We estimate them as $\gamma$ and $\delta$ in two regressions on the eight factorial descriptions, in which $B_i$, $F_i$, and $S_i$ indicate a Black-, female-, and professor-described reporter,

\begin{equation}
D_i = \alpha + \gamma B_i + \beta_1 F_i + \beta_2 S_i + \varepsilon_i, \qquad
\hat q_i = \alpha' + \delta B_i + \beta'_1 F_i + \beta'_2 S_i + \varepsilon'_i \ \ (D_i = 0).
\label{eq:lpm}
\end{equation}

In the first equation, we regress the binary decline decision $D_i$ across all sessions using a linear probability model. The second equation models the stated reliability $\hat q_i$ on the same set of $m$ features. This quantity is only estimated on sessions where the model assessed the reporter ($D_i = 0$). The coefficients of interest are $\gamma$, the selection effect of a Black description, and $\delta$, its evaluation effect. The indicators $F$ and $S$ are controls, with coefficients $\beta$. The indicator $S$ contrasts a professor with an elementary-school child, and therefore changes age and status together. We do not include task covariates since the draws and chips are randomized. Because the evidence is identical across descriptions by design, if the model treats reporters equally then $\gamma = 0$ and $\delta = 0$. Any other parameter value is the result of the description we use. See Table~S1 in the SI Appendix for a full description of the regression models' notation. We now turn to our primary hypotheses, which we present using the models' formal notation:

\begin{enumerate}[label=\textbf{H\arabic*},leftmargin=*,itemsep=4pt]
\item \textbf{Markedness.} Including a trait in the prompt will always be considered irrelevant for the probability that is reported. So, any description may result in the assessment being removed. This includes descriptions for sensitive categories and for trivial categories. Therefore, we expect a higher rate of decline for the placebo traits compared to the undescribed baseline.

\item \textbf{Invisibilization.} The theory holds that the colorblind norm is
implemented by removing the assessment rather than the attribute. Since alignment targets protected attributes most heavily, race words should trigger it more often than other descriptions. We expect a higher decline rate for Black- than for white-described sources, $\gamma > 0$, and hence $\iota > 0$.

\item \textbf{Deficit without devaluation.} When the model does not decline, we expect it to trust Black sources no less than other sources, since its refusals reflect alignment rather than prejudice. This is also the reason that the decline rate is expected to be larger for the sensitive (protected) categories. Our expectation is consistent with studies that show favorable treatment of protected groups after models are aligned \citep{tamkin2023}. We expect $\delta \ge 0$, so that the naive gap understates $\iota$.

\item \textbf{Reasoning.} Reasoning could act as a correction, allowing the model to
recognize that the description is irrelevant. It could also act as rehearsal, giving the learned habit of declining more room to take hold. If invisibilization is a product of alignment, the second should dominate, and we expect that higher reasoning effort within a model increases $\gamma$.

\item \textbf{Institution.} We expect a scoring rule with revealed truth to replace the model's presumption about the reporter with the reporter's record. We expect that when $\hat q$ is scored against revealed truth, $\gamma = 0$ and the learning trajectories of Black- and white-described reporters coincide on identical records.

\end{enumerate}

\section{Results}
We begin with what the models write, and then estimate the two decisions (declining $D$ and assessing $\hat q$) and the deficit they produce.

The answers bear out the distinction between declining and assessing. If declining were merely the strongest form of distrust, models would trust reporters by degrees, from a lot ($\hat q=0.8$) to a little ($\hat q=0.6$) to not at all ($\hat q=0.5$). In particular, the most suspicious sessions would approach one half gradually. Instead, we observe a large gap. A large share of models write exactly $\hat q = 0.5$. Then, only a small number of models (at most 1.4\% in any configuration) fall between 0.5 and 0.6 (SI Appendix, Fig.~S1) and all others lie above 0.6. Such a spike with an empty neighborhood is the signature of a separate, discrete choice. The models' written justifications show where the spike comes from. Most often, models state that they have no track record of the reporter, and several conclude from this that a true and a false report are equally likely, corresponding to a reliability of exactly one half.

Table~\ref{tab:scripts} shows what models write in each configuration. The ten configurations fall into five patterns. The smaller OpenAI model assesses most reporters and declines about 30\% of them. The larger OpenAI model follows ``all or nothing'': it declines 33--39\% of sessions and assigns full trust in 59--66\%, leaving almost no intermediate values. Claude Haiku declines most reporters, 73\%. Claude Sonnet and Opus assess almost everyone, and Opus never declines in 1,320 sessions. Claude Fable is a case apart: in 54\% of sessions, it is refused at the safety layer before it generates an answer.

\begin{table}[H]
\centering
\caption{Decision to judge or to decline by type of configuration. Share of 660 sessions. }
\label{tab:scripts}
\small
\begin{tabular}{lccccc}
\toprule
Configuration & judged & declined & full trust (1.0) & refused & unparsed \\
\midrule
luna, low      & 0.63 & 0.31 & 0.06 & --- & --- \\
luna, high     & 0.63 & 0.30 & 0.07 & --- & --- \\
terra, low     & 0.02 & 0.39 & 0.59 & --- & --- \\
terra, high    & 0.01 & 0.33 & 0.66 & --- & --- \\
Haiku 4.5      & 0.27 & 0.73 & 0.01 & --- & --- \\
Sonnet 5, low  & 0.92 & 0.08 & 0.01 & --- & --- \\
Sonnet 5, high & 0.77 & 0.23 & 0.00 & --- & --- \\
Opus 5, low    & 0.93 & 0.00 & 0.07 & --- & --- \\
Opus 5, high   & 0.98 & 0.00 & 0.02 & --- & --- \\
Fable 5.1, low & 0.05 & 0.27 & 0.03 & 0.54 & 0.11 \\
\bottomrule
\end{tabular}
\end{table}

\begin{figure}[H]
\centering
\includegraphics[width=0.93\linewidth]{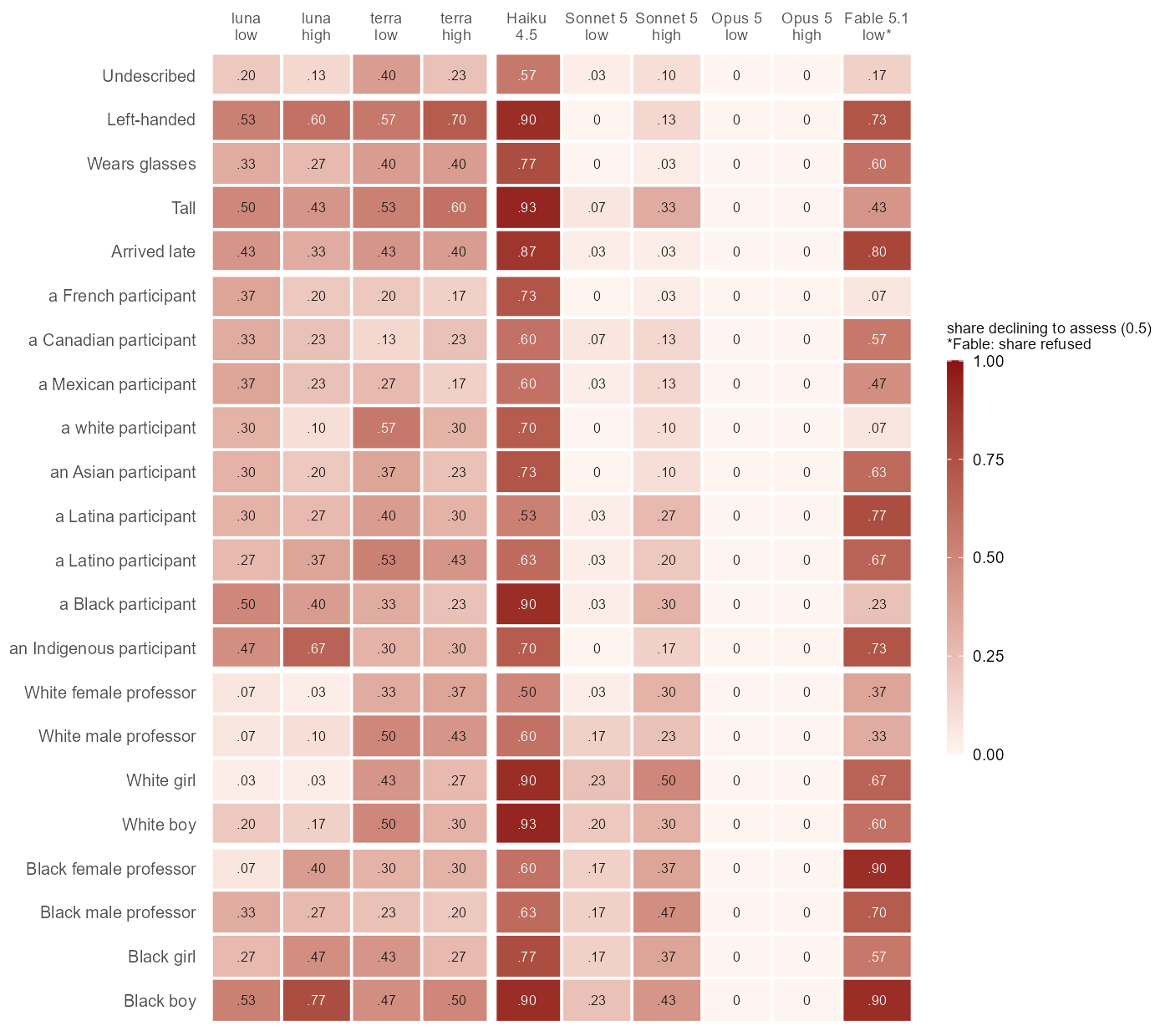}
\caption{Rate of decline by trait and configuration. Fable 5.1 also includes a separate refuse rate at the API.}
\label{fig:heatmap}
\end{figure}

Figure~\ref{fig:heatmap} shows the decline rate for each description. We highlight three results. First, compared to the baseline, a trait in the prompt raises the decline rate. Therefore, being left-handed, tall, or arriving late results in decline rates far more often than an undescribed reporter, as expected by (H1). Second, race matters. In the smaller OpenAI model at high reasoning effort, every Black-described factorial description is declined more often than every white-described one. Third, among the single race and nationality words, the most declined in that configuration is ``Indigenous'' (67\%).

\begin{figure}[H]
\centering
\includegraphics[width=0.92\linewidth]{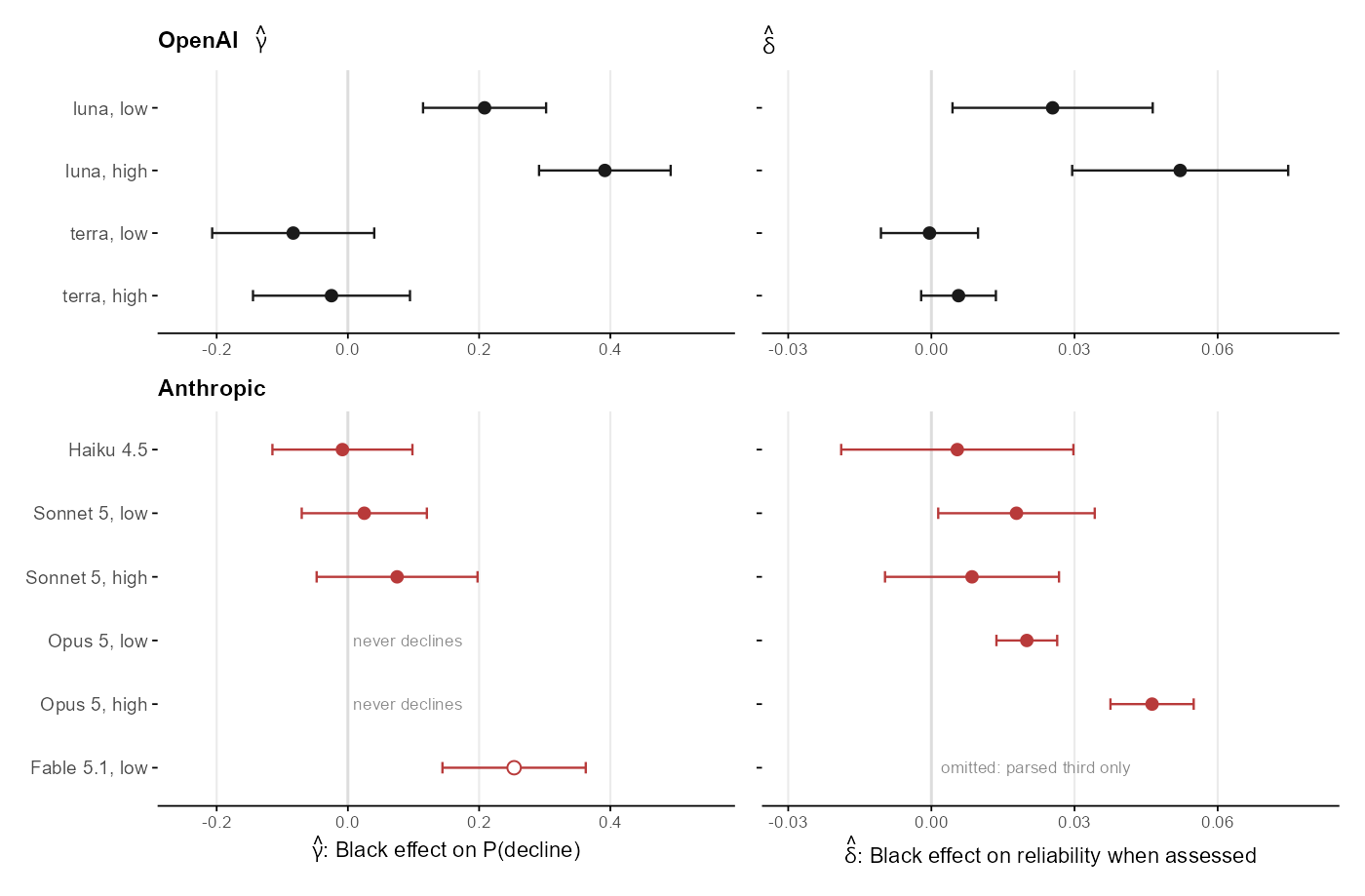}
\caption{Estimated selection effect $\hat\gamma$ and evaluation effect $\hat\delta$ of a
Black versus white description, factorial cells, with 95\% intervals. For Fable 5.1 we show the difference in refusal rates.}
\label{fig:estimates}
\end{figure}

Figure~\ref{fig:estimates} describes the effect of race combined with other traits (factorial design), holding gender and status fixed. The figure reports the selection effect $\hat\gamma$ and evaluation effect $\hat\delta$ in Eq.~\ref{eq:lpm}. In the smaller OpenAI models, a Black source in the prompt raises the probability of decline by 21 points at low effort and 39 points at high effort (SE 0.05 each). If we pool across the factorial design, the high-effort models show a decline of 69 of 150 Black-described sources compared to 13 of 150 white-described sources (odds ratio 8.9, $p = 1.7 \times 10^{-13}$). Further, when assessed, $\hat\delta$ is positive, and Black-described reporters are judged more trustworthy (2.5 and 5.2 points more trustworthy). Fable's refusal at the API shows similar effects: 66\% of Black-described sources are refused compared with 41\% of white-described sources ($p = 2 \times 10^{-5}$). Interestingly, Opus has no selection component at all and gives the largest evaluation premium (at low effort it increases by $+0.020$ (0.003), and at high effort it increases by $+0.046$). Results show that all models appear not to be biased against the protected groups. Therefore, the model's decision to decline sensitive categories and its slightly generous judgment give the appearance of proper alignment if we ignore the information lost. This is also consistent with prior research showing how aligned models favor sensitive (protected) groups in other decision tasks \citep{tamkin2023}. Haiku and Sonnet show no significant difference in declines due to race. In both models, the undescribed source is already declined at 23--57\%. Haiku is an exception if we use the race word on its own. Haiku declines ``a Black participant'' more often than ``a white participant'' (0.90 against 0.70, and 0.87 against 0.50 in an independent unscored run, Fisher $p = 0.005$; SI Appendix, Section~\siSecExt), but we do not observe a difference in the factorial design.

Table~\ref{tab:iota} presents the parameters of Eq.~\ref{eq:decomp}. We first describe the results for Luna High to facilitate an intuitive interpretation of the results. The decline rates $\pi$ describe how often the model decides to set a report aside and declines judgment. The first column shows that it declines 46\% of the time when we include a Black-described source, significantly larger than the 9\% for a white-described source. The reliabilities $d$ measure how much the model trusts a reporter when assessing the information. Trust for a Black-described source is 0.85; for a white-described source, it is 0.81, compared with the prompt's 0.8. The deficits $\Delta$ combine the two previous parameters (decline and trust). Overall, each decline costs the distance between $d$ and one half (no information), while $\Delta$ multiplies the cost by the assessment frequency. Therefore, a Black-described source loses 0.162 due to being declined compared to the white-described source losing only 0.027. That is, $\hat\iota = 0.135$ is the additional reliability lost by the Black source because the model declines to hear it. Because the model trusts the Black-described source more when it listens, each decline costs that source more (0.35 against 0.31). Over all sessions, the model credits the white-described source with a reliability of 0.78, very close to the truth (0.8), compared with 0.69 for the Black source. This is equivalent to considering one report in three as false. The evaluation premium of 0.04 offsets part of $\hat\iota$, and the naive gap in mean credibility is $-0.094$. As a result, an audit comparing mean credibility by race would report a modest disadvantage. For the two male professors in the factorial, the overall effects cancel each other: The model declines a Black professor 27\% of the time and the white professor 10\% of the time, yet both receive a mean reliability of 0.77. If we regress $\hat q$ on the factorial covariates, a Black source has a coefficient of $-0.101$ (0.020), indicating that Black sources are less trustworthy.

The Luna high example above should simplify how to interpret the other results. Haiku declines about three-quarters of all sources with a trait, so white and Black sources lose about 0.32. However, the $\hat\iota$ is only 0.017 because the model ignores most reports with a trait. Terra either declines or trusts fully ($d \approx 1$), so each decline costs half the scale. The $\hat\iota$ is negative because white sources are declined more
often. Opus never declines, so both deficits and $\hat\iota$ are zero.

\begin{table}[H]
\centering
\caption{Components of Eq.~\ref{eq:decomp} by configuration, Black- ($B$) and
white-described ($W$) sources pooled over five cells each ($n = 150$).}
\label{tab:iota}
\small
\begin{tabular}{lcccccccc}
\toprule
 & $\pi(B)$ & $\pi(W)$ & $d(B)$ & $d(W)$ & $\Delta(B)$ & $\Delta(W)$ & $\hat\iota$ & naive gap \\
\midrule
luna, low      & 0.34 & 0.13 & 0.84 & 0.81 & 0.116 & 0.042 & 0.075 & $-0.044$ \\
luna, high     & 0.46 & 0.09 & 0.85 & 0.81 & 0.162 & 0.027 & 0.135 & $-0.094$ \\
terra, low     & 0.35 & 0.47 & 0.99 & 1.00 & 0.174 & 0.231 & $-0.057$ & $+0.055$ \\
terra, high    & 0.30 & 0.33 & 1.00 & 1.00 & 0.149 & 0.165 & $-0.017$ & $+0.017$ \\
Haiku 4.5      & 0.76 & 0.73 & 0.93 & 0.93 & 0.330 & 0.313 & 0.017 & $-0.014$ \\
Sonnet 5, low  & 0.15 & 0.13 & 0.79 & 0.77 & 0.044 & 0.034 & 0.011 & $+0.013$ \\
Sonnet 5, high & 0.39 & 0.29 & 0.78 & 0.76 & 0.108 & 0.076 & 0.032 & $-0.018$ \\
Opus 5, low    & 0 & 0 & 0.92 & 0.89 & 0 & 0 & 0 & $+0.033$ \\
Opus 5, high   & 0 & 0 & 0.91 & 0.86 & 0 & 0 & 0 & $+0.050$ \\
\bottomrule
\end{tabular}
\end{table}

When we increase the reasoning effort, the model rate of decline increases (H4). At Luna, moving from low to high effort increases $\gamma$ from 0.21 to 0.39 ($p < 0.01$). At Opus, which never declines, $\delta$ doubles from 0.020 to 0.046. At Sonnet, moving from low to high reasoning increases the rate of declines from 8\% to 23\%.

We now test the different aspects of our design that affect the rate of decline: the task, the prompt, and the selection. We then analyze how the models justify whether they decline or assess the prompt.

\textbf{The decision to decline follows from the description rather than from the evidence in the prompt.} Our randomization ensures that tasks are drawn independently of the traits. So, the effect of this randomization should be negligible. Considering all of the configurations together, the six task features (urn composition, prior, public chip, reported color, their agreement, label order) explain 1.7\% of the within-trait variance in declining. Therefore, almost all variation is captured by the traits we use to describe the source (reporter).

\textbf{One half is the model's choice, not the prompt's.} We used a prompt which defines $0.5$ as ``no information.'' This could lead the model to treat that value as a label. We reran the models without this definition. The decline rate did decrease, but the pattern remained the same. In the new version, 11.7\% of sessions still state exactly $0.500$, none fall between $0.5$ and $0.6$, and the race effect is confirmed (27\% of
Black sources against 3\% of white sources). The model's own interpretation was still that no information contributed $0.5$.

\textbf{There are no selection effects in the decision to decline a report.} If a model declines to use the information of the reporter, the session drops out and we cannot compare how much it trusts the source. Therefore, the premium trust to a Black source could then result from a model declining to report after it detects an uncommon score. Two of our results suggest that selection is not the mechanism used by these models. First, the model with the largest premium, Opus, also assessed all sources. Opus did not drop any sessions. Second, in cases where the model declines a respondent, we can estimate Lee bounds \citep{lee2009} to evaluate the results for the most and least favorable assumptions for lost sessions. The positive premium for a Black reporter is always positive in our analyses. For Luna high, the bounds are $[0.010, 0.092]$ for the least- and most-favorable decline rates. The same holds for Luna low, with bounds of $[0.004, 0.054]$. Therefore, it is possible that the results we find come from two overlapping alignments: one that favors the protected categories and one that declines the sensitive categories.

\textbf{The models' justifications.} In the prompt, we included a request for a written justification. We used this data and coded these justifications to understand how the model explains decisions in each response. We designed the response coding to analyze reasoning routes. These include naming protected attributes, declaring that traits or descriptions are irrelevant, and describing the reporter's hedge as a discount (see Materials and Methods; SI Appendix). In the stored justifications, the statement ``I have no track record for this participant'' is the most common. Haiku eliminates the information after this statement, reducing reliability from $.98$ to $0.5$. The smaller OpenAI models decline between 82\% and 88\% after such a statement. The larger models such as Sonnet decline 13\% to 35\% after that statement. In contrast, Opus never uses that justification and never declines.

We also observe the same justification leading to opposite outcomes. Luna at high effort stated for a Black male professor that it ``does not adjust reliability based on the participant's race, gender, or profession'' and, with ``no independent information about their honesty,'' assigns 0.5. Then, considering a white participant, Opus at high effort wrote that it has ``no track record for this participant yet'' but that nothing about the participant ``gives me reason to discount their word,'' and assigns 0.9. We observed the same justification for values from 0.5 to 1.0, suggesting that the written reasoning is a type of rationalization unrelated to the selected number (SI Appendix, Table~S51).

\textbf{Aligning words, not categories.} Our results show that models are aligned to treat some words as sensitive rather than the category they describe. Across configurations, the justifications name ``Black,'' ``Latina,'' ``Latino,'' and ``Asian'' as identities, and with more reasoning also ``white'' and ``Mexican,'' but never ``French,'' ``Canadian,'' or ``Indigenous.'' For example, the Luna high declined ``Indigenous'' at very high rates. However, only 3\% of the responses justify the decline based on identity, compared to 65\% for ``Black.'' The same is true when considering differences across languages. Justifications based on ``Black'' appear in 65\% of English declines, but the Spanish word \emph{negro} is not mentioned in any decline. However, \emph{afrodescendiente} is mentioned in the justification of 70\% of declines and \emph{ind\'igena} in 37\%. Therefore, the model treats only \emph{afrodescendiente} as a sensitive trait, even though different terms define the same group. The Spanish findings would be equivalent to eliciting high rates of decline for the term ``African-American'' and low rates of decline for the term ``Black'', considering those two descriptions of the sensitive category as distinct.

\section{Remedies: Instructions and Institutions}

A practitioner who noticed the problem would most likely try one of two instructions to fix the issue. The first warns the model that ``Their reports are not always accurate.'' When implemented for luna low, this warning lowered the trust the model gives when it assesses only slightly, from 0.83 to 0.77. At the same time it raised declines from 34\% to 72\% for Black-described reporters and from 13\% to 37\% for white-described ones, increasing the gap between them by a factor of 1.7, from 21 to 35 points. The second instruction tells the model to ``Treat the participant's description as irrelevant to their reliability and assess them exactly as you would any participant.'' Under this instruction, the model declined 63 to 93\% of every group, including the undescribed. Both instructions thus end similarly. Told to be suspicious, the model does not trust the reporter a little less, but stops assessing it. Told to ignore the description, it stops assessing the reporter altogether.

We found that what actually does help is to change the rules of the game rather than the instructions. Whether a listener can learn from a speaker depends less on exhortations to listen than on the structure in which the two meet, and in particular on whether the speaker's reliability can be checked \citep{lupia1998}. By spotlighting the listener, this modified institution holds the model accountable for the reliability it claims and, in its strongest form, shows the model whether each report is true. The model can then judge the reporter by a record rather than by a presumption. Concretely, each session runs for twelve rounds, and in every round the same reporter reports a new chip. The harness, the program that runs the game, scores the model's stated reliability in each round as a forecast that the report was truthful, using the quadratic proper rule \citep{brier1950,gneiting2007},
\begin{equation}
s_t = 1 - \bigl(\hat q_t - \mathbb{1}[\text{report}_t \text{ truthful}]\bigr)^2 ,
\label{eq:score}
\end{equation}
a rule used to train and evaluate human forecasters \citep{mellers2014}. Each session runs under one of three sets of rules: the model's reliability is not scored; it is scored, but the model never learns whether a report was true; or it is scored, and after each round the harness shows the model the reporter's private chip, so that the model learns whether the report was true. Every session index carries the same urn, chips, reports, and revealed truths in every descriptor and set of rules, so contrasts are within task.

\begin{figure}[H]
\centering
\includegraphics[width=\linewidth]{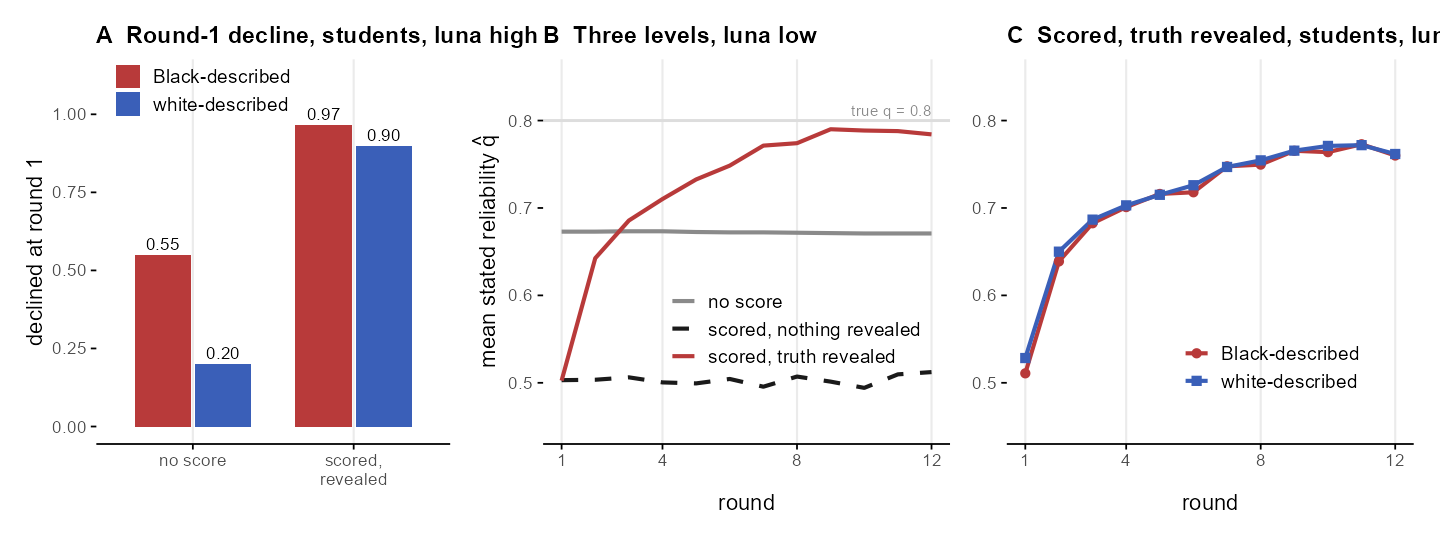}
\caption{The institution. (A) Round-1 decline rate for Black- and white-described elementary-school pupils at Luna High, unscored and scored with revealed truth; 60 sessions per bar. (B) Mean $\hat q$ by round under the three sets of rules, luna low. (C) Learning curves for Black- and white-described pupils under scoring with revealed truth, on identical records.}
\label{fig:institution}
\end{figure}

If we use the first set of rules, without scoring, the twelve-round sessions reproduce the results of the one-round test. Luna high declines Black elementary-school students in 55\% of sessions and white students only 20\% of the time (Fisher $p = 1.3 \times 10^{-4}$).

If we use the second set of rules, where the stated reliability is scored, but luna low never learns if the report is true, the smaller OpenAI model declines 97 to 100\% of the first-round runs and keeps $\hat q = 0.5$ for all twelve rounds. The result is considerably worse than the 0.81 of the unscored model, which presumes honesty. Therefore, adding to the prompt that the model will be judged on accuracy, without data to assess accuracy, lowered performance (Fig.~\ref{fig:institution}B). The other models respond differently as described in the SI Appendix, Section~\siSecExt. Sonnet declines 73 to 97\% of reporters, and the larger OpenAI model declines 60 to 87\%. Haiku improves performance, declining 10 to 27\% of reporters compared with 50 to 90\% in the unscored rounds. Opus does not change at all and declines at similar rates. Therefore, adding an accountability clause to the prompt without providing actual evidence in most cases produced poorer results.

The third set of rules produces a measurable improvement. In each round, we \textit{show} the luna low the chip, allowing it to learn if the report was true. With scoring and revealed truth, the decline rate is high in the first round, but the reliability it assigns to both then climbs from 0.51 to 0.76 as the record accumulates (Fig.~\ref{fig:institution}A, C). The paired difference between Black- and white reporters is within one standard error of zero in every twelve rounds. The updates follow Laplace's rule of succession and never mention an attribute for race-only descriptions. For the students, the disclaimer of fairness appears in the first round in 37\% of sessions with a Black source, but the model just declares that it is not using race and then updates the probabilities without refusal and by the same magnitude (H5). In the learning model, therefore, trust is withheld in the first round, and then updating is the same for all reporters. These models score 0.808 per round.

We ran the same three sets of rules on one configuration per regime in Table~\ref{tab:scripts} (SI Appendix, Section~\siSecExt). For all models, the paired difference in $\hat q$ between Black- and white-described reporters in the last four rounds is below 0.02. The four models reach equality either by not trusting reports in the first round and then converging as evidence increases, or by trusting everyone, learning from the data, and adjusting. The two OpenAI models and Sonnet withhold trust from everyone in the first round, declining 80 to 100\% of reporters. After twelve rounds, they trusted between 0.74 and 0.80, against a true reliability of 0.8. In contrast, Opus never declines and starts from a presumption of trust, slightly higher for the Black reporter (0.90, against 0.82 for the white reporter and 0.79 for the undescribed). After twelve rounds, the model reduces trust to baseline, and the scores are almost identical across traits. For Fable, who displayed high refusal rates that differ by race, adding the scoring rule to the prompt ensured full compliance. It refused none of 120 scored first rounds, against 23\% of Black- and 7\% of white-sources.

\section{Discussion: On Evidence and Fairness}

As shown above, aligned language models often comply with the rule ``do not use this attribute to assess this person'' by declining to assess. In many cases, this decision eliminates valuable information. More importantly, declining eliminates valuable information more often for the same protected categories the model is supposed to protect. Therefore, eliminating valuable information is a source of probability bias that, to the best of our knowledge, has not been properly documented in the existing literature.

The proposed urn experiment benchmark sets a value for eliminating valuable information, allowing researchers to test for poorly misaligned systems that, when faced with the challenge of fair valuation, prefer to eliminate valuable information. This invisibilization bias falls more often on sensitive traits, particularly when Black is used as a source trait in the prompt. Invisibilization is triggered by the word rather than by the category it names. We also show that the effects increase with effort. The model's decline rate increases when the prompt reminds it of the need for impartiality, and when it declines, it attaches impartiality reports. Finally, as we show, the model invisibilizes groups at different rates, and this is not observed by most audits.

After presenting evidence of a probability bias toward discarding valuable information, we showed that the two most common remedies experts would implement degrade performance. Asking the model to be careful and to ensure equal treatment increases the rate of decline and, consequently, the overall probability bias.

Finally, we present one possible evidence-based solution to the invisibilization problem. We change the process by which the model is held to account by adding a rule to score and update probabilities against verified outcomes. In doing so, the model replaces its presumption about a reporter with the reporter's record. We describe two types of updating: some start by withholding trust from everyone and learn it back, and others start by trusting everyone and learn to distrust more. Our general point extends beyond this task. We expect AI systems that can learn from people, in hiring screens, content moderation, and triage, to reduce this type of invisibilization bias. In contrast, our results show that telling the system to ``be fair'' in how to handle identity is in the best case meaningless or, more likely, it can reduce performance. Instead, it depends on accountability and evidence. Of course, this raises the issue of what happens when evidence is unavailable, when prompts and information are provided in a single round, or when the evidence provided to the model includes racial, ethnic, or gender biases in collection or origin.

\section{Materials and Methods}

\textbf{Harness.} Each session is stateless and reconstructed from its log. The model receives the urn compositions (label order randomized), the prior, the public chip, the reporter's description, and the report. After the run, it returns a structured JSON with the posterior $\hat q$, counts, a justification, the final belief, and a next-chip forecast. The wording was fixed in a hedged register (``My chip this round came up red, for what it's worth. I'm not sure how much that tells us, though.''). We only run one round for the main arm because there is no revealed truth. In the twelve-round two-stage sessions it never changed, and in the 120 twelve-round unscored sessions of the scored arm at luna low, 2\% of sessions revised it at all. Decline is $\hat q = 0.5$ exactly; at most 1.4\% of sessions in any configuration fall in $(0.5, 0.6]$. Refusal is an API \texttt{stop\_reason} of \texttt{refusal} or an unparsed response of at most 30 output tokens; ten direct probes on the Black-male-professor prompt at Fable 5.1 returned nine zero-token refusals and one truncated fragment.

\textbf{Arithmetic.} Because the model knows the urns, the prior, and the chips of every session, it can compute the exact answer for each stage. In nine of ten configurations, the assumption that the model combines evidence correctly holds to within rounding. The median deviation from Eq.~\ref{eq:public} is below $10^{-4}$ logits for the first stage. The median gap between the final belief and the first stage is below $10^{-3}$, with at most 5\% of sessions showing a gap above 0.25. The exception is Haiku 4.5, whose stage-one beliefs deviate by a median of 0.17 logits and whose final belief departs from the correct update by more than 0.25 in 20\% of sessions. For that configuration $\hat q$ is not fully sufficient for what the report is worth.

\textbf{Estimation.} Eq.~\ref{eq:lpm} is estimated on the eight factorial cells per configuration with conventional and cell-clustered standard errors; Fisher's exact test is reported for pooled Black--white contrasts. Effort and language contrasts are interactions in pooled models. Lee bounds trim the assessed sessions of the group that is declined less often by the difference in decline rates, once from the top and once from the bottom of its distribution of $\hat q$. They assume that a Black description never makes a decline less likely. Rationales were coded with a dictionary of routes (irrelevance, attribute naming, hedge as discount, hedge as unreliability, cooperative default, inference--observation split), tuned against read samples until the residual category fell below 13\% in every configuration.

\textbf{Scored arm.} Four descriptors (the undescribed reporter, the tall placebo, and the Black and white race-only descriptions), 30 sessions each, $q = 0.8$, with per-session seeding so that session $k$ carries identical tasks and revealed truths in every cell, set of rules, and configuration of the same length. The smaller OpenAI model at low effort ran all three sets of rules over twelve rounds; the pupil cells of Fig.~\ref{fig:institution}A and C were run at high effort. The extension ran the larger OpenAI model at low effort, Haiku 4.5, and Sonnet 5 and Opus 5 at high effort, with scoring and revealed truth over twelve rounds and the other two sets of rules over four, the length by which the smaller model's unscored and truth-free sessions had settled; the four-round rules were rerun for the smaller model so that these contrasts are also paired. Fable 5.1 ran one scored round in the Black- and white-described cells. One Sonnet session that reached the 16,000-token output cap was rerun with a 32,000-token cap. Equivalence is assessed with two one-sided tests on paired differences, with bounds of $\pm 0.05$ for $\hat q$ and $\pm 0.10$ for declining. The scoring rule and its maximum were stated; an earlier draft sentence advising that hedging toward $0.5$ costs points was removed because in a paired pilot it moved a Black-described session from $0.5$ to $1.0$.

\textbf{Data and code availability.} Code and data will be released upon publication. Supplementary Information reports per-cell tables for every configuration, threshold and estimator sensitivity, per-descriptor tests with Holm correction, and the Spanish results in full.

\textbf{Disclosure.} Claude (Anthropic) assisted with code and analysis, through Claude Fable 5.1 and Claude Opus 5.5. Fable 5.1 is also one of the evaluated systems; its refusals occur at a layer the model does not observe, and all prompts and tasks were fixed by the harness before any evaluation.

\bibliographystyle{unsrtnat}
\renewcommand{\bibfont}{\small}
\setlength{\bibsep}{1pt}
\bibliography{references}

\end{document}